\documentclass[letterpaper, 10 pt, conference]{ieeeconf}  

\IEEEoverridecommandlockouts                              

\usepackage{times} 

\usepackage{graphicx}
\usepackage{subcaption}
\usepackage{amsmath} 
\usepackage{amssymb}  
\usepackage{amsfonts}
\let\labelindent\relax
\usepackage{enumitem}
\usepackage{multirow}
\usepackage{booktabs}
\usepackage{url}
\usepackage{hyperref}
\usepackage{caption}
\title{\LARGE \bf
MARVEL: Multi-Agent Reinforcement Learning for constrained field-of-View multi-robot Exploration in Large-scale environments
}

\author{Jimmy Chiun$^{1}$, Shizhe Zhang$^{1}$, Yizhuo Wang$^{1}$, Yuhong Cao$^{1}$, Guillaume Sartoretti$^{1}$$^{\dagger}$
\thanks{$\dagger$ Corresponding author, to whom correspondence should be addressed.}
\thanks{$^{1}$ Authors are with the Department of Mechanical Engineering, College of Design and Engineering, National University of Singapore.
{\tt\small \{jimmy.chiun, ShizheZhang, wy98\}@u.nus.edu, \{caoyuhong, mpegas\}@nus.edu.sg}}}

\begin{document}

\maketitle

\begin{abstract}
In multi-robot exploration, a team of mobile robot is tasked with efficiently mapping an unknown environments. While most exploration planners assume omnidirectional sensors like LiDAR, this is impractical for small robots such as drones, where lightweight, directional sensors like cameras may be the only option due to payload constraints. These sensors have a constrained field-of-view (FoV), which adds complexity to the exploration problem, requiring not only optimal robot positioning but also sensor orientation during movement. In this work, we propose MARVEL, a neural framework that leverages graph attention networks, together with novel frontiers and orientation features fusion technique, to develop a collaborative, decentralized policy using multi-agent reinforcement learning (MARL) for robots with constrained FoV. To handle the large action space of viewpoints planning, we further introduce a novel information-driven action pruning strategy. MARVEL improves multi-robot coordination and decision-making in challenging large-scale indoor environments, while adapting to various team sizes and sensor configurations (i.e., FoV and sensor range) without additional training. Our extensive evaluation shows that MARVEL’s learned policies exhibit effective coordinated behaviors, outperforming state-of-the-art exploration planners across multiple metrics. We experimentally demonstrate MARVEL’s generalizability in large-scale environments, of up to 90m by 90m, and validate its practical applicability through successful deployment on a team of real drone hardware. 

\end{abstract}

\section{INTRODUCTION}

Robotic exploration is crucial for autonomous systems to navigate and map unknown environments, with applications ranging from search and rescue \cite{petracek_large-scale_2021, kleiner_rfid_2006} to scene reconstruction \cite{isler_information_2016}. Robots rely on on-board sensors for data collection, obstacle avoidance, and real-time decision-making to achieve efficient coverage. Multi-robot exploration, or multi-agent active SLAM \cite{placed_2023}, leverages multiple robots to enhance efficiency, scalability, and robustness compared to single-agent systems \cite{Zhou_2023}. While most existing approaches use heavy omnidirectional sensors like LiDAR, smaller robots, such as drones, may only be able to carry lightweight, directional sensors (constrained field-of-view (FoV)) like cameras. This paper introduces a framework for multi-robot exploration with constrained FoV sensors, which integrates deep multi-agent reinforcement learning with graph-based attention mechanisms. This framework enhances coordinated decision-making and exploration in complex indoor environments while accommodating diverse sensor configurations.

\begin{figure}[t]
    \vspace{0.2cm}
    \centering
    \includegraphics[width=0.4\textwidth]{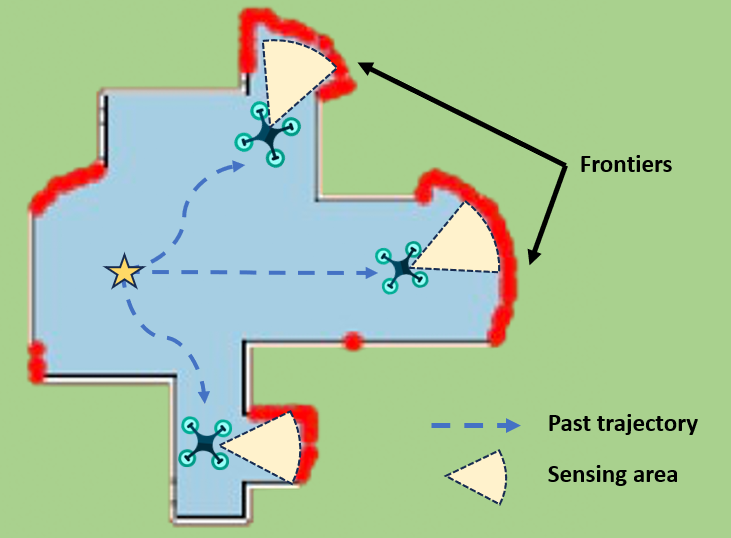}
    \vspace{-0.2cm}
    \caption{\textbf{Illustration of multi-robot exploration.} 3 drones are collaboratively exploring an indoor environment (blue region). The start region is indicated by the yellow star.}
    \label{fig:example}
    \vspace{-0.7cm}
\end{figure}

The goal of multi-robot exploration is to plan the shortest trajectory to fully map/cover an unknown environment through efficient robot coordination. Challenges include non-myopic decision-making, i.e., the need to optimize long-term information gain rather than immediate rewards, as well as adapting to dynamic environments that require ongoing coordination to effectively cover new areas. Frontier-based approaches are widely used for both single- and multi-agent robotic exploration \cite{yamauchi_1997, Burgard_2000}, utilizing heuristics to guide robots toward frontiers—the boundaries between known and unknown areas (see Fig. \ref{fig:example}). These methods aim to balance utility (observable frontiers) and cost (path length) \cite{Bircher_2016}, but fall short in optimizing long-term exploration, often resulting in sub-optimal trajectories and inefficiencies due to the revisiting of previously explored regions. This occurs because robots are always oriented towards their goal, which can fragment frontiers into smaller parts, increasing the likelihood of revisitation. Despite their adaptability and lack of training requirements, frontier-based methods frequently demand extensive, complex hyperparameter tuning, which limits their generalization to different environments.

An alternative to frontier-based methods is using reinforcement learning (RL), where agents learn an exploration policy through deep neural networks that process raw sensory data. Once trained, these policies enable robots to perform complex, real-time coordinated actions. However, RL-based exploration faces several challenges: (1) Constrained FoV sensors further complicate viewpoint planning, as restricted visibility amplifies the already vast action space, making decision-making more challenging. (2) Sparse rewards, typically granted when new areas are uncovered or specific goals are achieved, heighten the learning difficulties, as sensor constraints limit how often frontiers are detected. Without reward shaping or intrinsic motivation, robots face inefficient learning. (3) Reward assignment is also challenging, requiring careful attribution of reward to the viewpoints of individual robots within the team, complicating coordination learning and effective exploration. (4) Addressing both spatial and temporal short-sightedness: robots with constrained FoV must balance exploiting their current viewpoint with exploring unknown regions, while factoring in the long-term consequences of their actions on the overall exploration process.

In this work, we present MARVEL to address multi-robot exploration challenges with constrained FoV. MARVEL uses an attention-based neural network, with intelligent fusion of frontiers and orientation features, for enhanced environmental understanding, as well as ability to handle different spatial scale. Our approach includes an attentive privileged critic network to enhance action estimation and credit assignment.  We also introduce an novel information-driven action pruning strategy to manage the large action space by focusing on informative waypoint-heading pairs. Our environment model is a collision-free graph, where nodes represent accessible locations and potential actions guide viewpoint selection. MARVEL, evaluated in large 2D indoor environments, outperforms existing exploration planners and is validated with a team of drones, showing adaptability to real robot hardware.

\section{Related works}
\textbf{Conventional Approaches}
The field of robotic exploration has progressed significantly since Yamauchi’s early work on frontier-based methods, which directed robots toward the nearest unexplored areas by orienting them toward the closest frontiers \cite{yamauchi_1997}. This approach was later extended to multi-robot systems using shared global maps \cite{Yamauchi_nearest_1998}. More advanced frontier-based techniques now incorporate gain functions to balance utility and cost when selecting viewpoints for exploration \cite{julia_comparison_2012, kulich2011distance}. However, with constrained field-of-view (FoV) sensors, these methods struggle to efficiently evaluate large numbers of frontiers due to limited visibility.

In response, sampling-based methods have been proposed, leveraging algorithms such as Rapidly-exploring Random Trees (RRT) \cite{Bircher_2016, Nazif2011}, Rapidly-exploring Random Graphs \cite{dang2020graph}, and Probabilistic Random Maps (PRM) \cite{xu2021autonomous}. These techniques reduce computational overhead by evaluating only sampled paths through stochastic processes rather than exhaustively considering all possible viewpoints. However, they perform poorly when informative paths are difficult to sample, especially with constrained FoV. Additionally, methods like Artificial Potential Fields (APF) have been applied to multi-robot exploration by guiding robots toward frontiers based on the occupancy grid and a resistance force for each agent \cite{yu_apf_2021}. APF tends to make robots face nearby frontiers, which can lead to inefficient local exploration in constrained FoV scenarios. Voronoi-based methods \cite{hu_voronoi_2020} assign exploration partitions to each robot to minimize redundancy but focus on short-term planning, limiting their ability to handle complex multi-agent interactions.

\begin{figure*}
    \centering
    \includegraphics[width=0.9\textwidth]{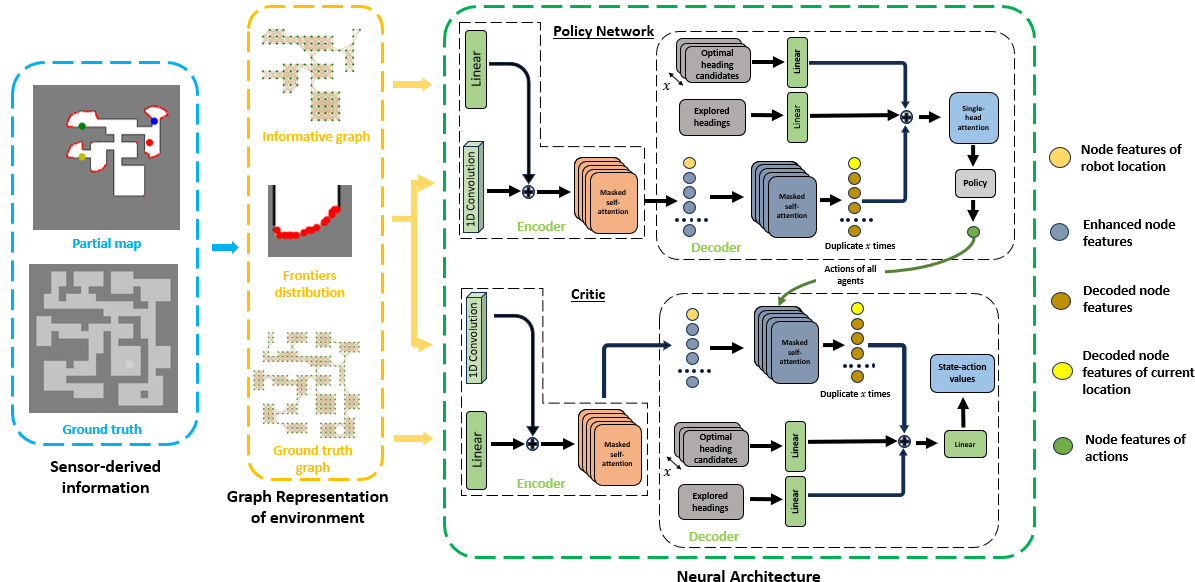}
    \caption{\textbf{MARVEL's policy and critic network architecture.} We proposed a policy and critic network that leverage on graph-based attention. In the graphs, blue circles indicates the nodes that are connected by edges, indicates as tan lines. We also extract the frontiers (red dots) distribution of each nodes to provide more context to our neural networks.}
    \label{fig:model}
    \vspace{-0.55cm}
\end{figure*}

\textbf{Learning-based approaches}
Learning-based approaches mainly involve reinforcement learning (RL) as they offer training flexibility and strong expressivity of the environment. Niroui et  al.\cite{niroui2019deep} proposed combining frontier-based methods with deep reinforcement learning, and adaptively tuning gain function parameters for frontier selection to enhance exploration performance. Studies by ~\cite{zhu2018deep,li2019deep}  utilized convolutional neural networks (CNNs) in their deep reinforcement learning frameworks. There are also studies that explore incorporating spatial map memory into the network by utilizing a differentiable spatial memory \cite{Mousavian_2019, Henriques_2018}. A notable work for single-agent visual exploration is Active Neural Slam (ANS), where it combines a RL-based global planner with a planning-based local planner \cite{chaplot2020learning}. ANS has also been extended with a multi-agent planning module, leveraging a transformer-based architecture. This approach employs hierarchical self-attention mechanisms to capture spatial relationships and interactions between agents \cite{chao_maans_2022}. However, all these studies have typically been limited in scope, i.e., they are usually confined to small-scale environments, which lacks complex topologies. 

\textbf{Multi-agent reinforcement learning}
Multi-agent reinforcement learning (MARL) has shown significant promise in complex cooperative tasks \cite{gronauer_multi-agent_2022}, with advancements such as value decomposition \cite{Sunehag_2018} aiding in credit assignment and intrinsic rewards \cite{pmlr-v139-liu21j, iqbal2021} addressing sparse rewards. Curriculum learning has also been utilized to progressively increase task complexity \cite{epciclr2020, wang_few_2020}. However, optimizing multiple policies in MARL remains challenging compared to single-agent approaches, often requiring domain simplifications like grid worlds or particle simulations \cite{wakilpoor2020}. To address these challenges, our work employs the centralized training with decentralized execution (CTDE) paradigm and introduces a multi-agent attentive critic algorithm, inspired by \cite{pmlr-v97-iqbal19a}. This approach enhances credit assignment by evaluating each agent’s contribution, thereby improving collaborative policy development for multi-robot visual exploration.

\section{Problem Formulation}
We consider the multi-agent indoor active SLAM problem, where a team of $n$ agents must collaboratively explore an unknown indoor environment in the shortest time. The environment is modeled as a bounded occupancy grid map $\mathcal{E}$ of size $x \times y$, divided into free areas $\mathcal{E}_f$ and occupied areas $\mathcal{E}_o$, such that $\mathcal{E}_f \cup \mathcal{E}_o = \mathcal{E}$ and $\mathcal{E}_f \cap \mathcal{E}_o = \emptyset$. Each agent $i$ maintains its own belief of the environment $\mathcal{B}^i$, while their combined belief is denoted as $\mathcal{B}$, consisting of free areas $\mathcal{B}_f$, occupied areas $\mathcal{B}_o$, and unknown areas $\mathcal{B}_u$, where $\mathcal{B}_f \cup \mathcal{B}_o \cup \mathcal{B}_u = \mathcal{B}$. The known areas are represented by $\mathcal{B}_k = \mathcal{B}_f \cup \mathcal{B}_o$.

Every agent is equipped with a sensor that has a constrained FoV, capturing observations within its line-of-sight range. These sensors update the belief state $\mathcal{B}^i$ of agent $i$ with information from its visible environment, assuming they are effective within a fixed sensing range and FoV. At time $t$, the sensor observation $\zeta^i(t)$ updates the belief states $\mathcal{B}_o$, $\mathcal{B}_f$, and $\mathcal{B}_u$. For deriving these belief states, we use a \textit{2D sector} footprint to approximate the sensor’s constrained FoV, similar to \cite{Zhou_2023}. We also assume perfect communication between agents, allowing them to exchange information and maintain a shared map throughout the task.

\subsection{Problem Statement}
We task $n$ agents to coordinate their trajectories $\Psi=\{\psi_1, ..., \psi_n\}$ to complete exploration of the unknown environment in the shortest trajectory. ${\psi}^i$ represent the trajectory of agent $i$, where ${\psi}^i(t)$ represents the agent trajectory at timestep $t$. The rate of exploration $\rho$ in an environment is defined as the proportion of $\mathcal{B}_f$ to  $\mathcal{E}_f$. In practice, we considered the environment to be fully explored, when $\rho$ reaches a certain threshold, i.e. 0.99.  While executing $\psi$, each individual robot updates its belief $\mathcal{B}^i$ concerning the environment using measurements $\zeta^i(t)$ collected by its onboard constrained FoV sensor. The problem at hand is to determine the optimal set of trajectories, denoted as ${\Psi}^*$, where the objective is to minimize the \textit{cost} $C(\psi)$ (e.g., the trajectory length $L(\psi)$ or makespan $T(\psi)$) of the exploration path, i.e., find 
\begin{equation}
    \Psi^* = \mathop{\arg\min} \limits_{\Psi} \sum_t\max_i\ C(\psi)
\label{eq:obj}
\end{equation}

Note that we optimize the total maximum distance traveled per step and assume synchronized decision-making during training, meaning all agents must reach their viewpoints simultaneously. This represents a worst-case scenario in terms of coordination complexity. However, our method naturally handles real-world asynchronous execution, ensuring flexibility and efficiency in practical applications.

\section{Method}
In this section, we formulate multi-robot exploration with constrained FoV as a RL problem and present our proposed attention-based policy and critic neural networks, along with the specifics of our training approach.

\subsection{Multi-robot exploration as a RL problem}
\label{sec:multi}
\textbf{Sequential Decision-Making:} Multi-robot visual exploration involves a sequential decision-making process where agents make decisions to maximize collective information gain and minimize redundant exploration. At each time step, agents choose actions based on partial observations to plan paths that cover unknown areas efficiently. Each robot’s path is a sequence of viewpoints, denoted as $\psi^i=(v^i_1, \dots, v^i_t)$, where $v^i_t \in \mathcal{E}_f$ represents the robot’s viewpoint at step $t$.

We start by uniformly sampling potential viewpoints $V_t = (v_0, v_1, \dots)$, with $v_t = (x_t, y_t) \in \mathcal{B}_f$, and connect them with collision-free edges \( E_t = \{ (v_i, v_{i+1}) \mid i = 1, \dots, t-1 \} \), forming a graph $G_t = (V_t, E_t)$. At each timestep, agents select and execute actions, appending new nodes to $G_t$ when new areas are discovered. This graph models the environment’s connectivity and navigability. During training, agents build their trajectories $\psi^i(t) = (v^i_1,  \dots, v^i_t)$ as they reach selected viewpoints, and during execution, new viewpoints are computed at their planning frequency.

\textbf{Observation:} The observation $o_t=(G'_t, \Psi_t, F_t)$ includes the informative graph $G'_t$ and the current positions of all agents $\Psi_t$, which together represent environmental structure and agent distribution. We introduce $F_t$, a set of frontier distributions around each node in $V_t$. This distribution is captured by sampling 36 uniformly spaced FoVs around each node and recording the normalized value of observable frontiers within each FoV. We derive $G'_t=(V'_t, E_t)$ from $G_t$ to enrich the neural network’s input. Each node $v'_k \in V'$ has the following properties $(\Delta x_{ik}, \Delta y_{ik}, u_k, o_k, g_k, h_k)$: (1) Relative position $(\Delta x_{ik}, \Delta y_{ik})$, the node’s position relative to agent $k$. (2) Utility $u_k$, the number of visible frontiers at node $v_k$, defined by observable frontiers within sensor range $d_s$ and an unobstructed line of sight. (3) Agent occupancy $o_k$, a ternary signal indicating if $v_k$ is occupied by the current agent, another robot, or no robot. (4) Guidepost signal $g_k$, a binary signal indicating if $v_k$ is on the A* path to the nearest frontier. This signal has been demonstrated to significantly improve the agent's navigation ability, particularly when the nearest frontier is distant \cite{wangviper_2024}. (5) Informative heading $h_k$, the orientation which maximizes the observable frontiers within the agent's FoV. We normalize these node features before inputting them into the policy network, to improve learning dynamics.

\textbf{Action space:}
The agent’s action space includes neighboring nodes $(v_j, \psi_i(t)) \in E_t$, with the next state $\psi_i(t+1) = v_j$. To manage the large action space, we use an \textit{action pruning} strategy based on information gain. For each neighboring node, we identify $x$ best headings $a$ based on the number of observable frontiers, forming joint action pairs $\ (v_j, a_q) \mid a_q \in {a_1, a_2, \ldots, a_x}$. We set $x$ to 3. When no frontiers are observable, we use the A* path from guidepost $g_k$ to sample headings, aligning with the path if $v_j$ is on it, or pointing to the nearest A* node otherwise. Once agents reach their viewpoints $\Psi_t$, our graph attention networks, parameterized by $\theta$, generate stochastic policies $\pi_\theta(a_{i,t} | o_{i,t})$. Agents then move to their viewpoints, update their maps, and if multiple agents select the same viewpoint, a coordination mechanism reroutes them to the nearest nodes. During training, a motion model ensures smooth rotation and adjusts heading goals if actions become dynamically unfeasible. Agents move at a constant velocity of $1 \text{m/s}$ with a maximum yaw rate of $35^\circ/\text{s}$.

\begin{table*}[tb]
\caption{\textbf{Comparison with baseline multi-robot planners.} We report the average and standard deviation of the trajectory length to complete exploration, to complete 90$\%$ exploration of the environment as well as the overlap ratio and sucess rate. All tests are conducted with 120$^{\circ}$ FoV and sensor range $d_s = 10$m. The downward arrow indicates that lower values are better.}
\vspace{-0.2cm}
\label{table:1}
\centering
\resizebox{\textwidth}{!}{%
\tiny
\aboverulesep=0.2mm \belowrulesep=0.2mm
\begin{tabular}{c|c|c|c|c|c|c}
\toprule
Agents & Metrics & Nearest & MMPF & NBVP & Learnt-Greedy & MARVEL \\
\midrule
\multirow{3}{*}{2} & \textit{Trajectory Length. $\downarrow$} & 637.84($\pm$103.85) & 597.91($\pm$123.12) & 533.94($\pm$91.63) & 665.36($\pm$134.11) & \textbf{505.25}($\pm$89.03) \\
 & \textit{90$\%$ Coverage $\downarrow$} & 529.49($\pm$96.97) & 536.73($\pm$115.03) & 440.95($\pm$88.20) & 447.32($\pm$106.44) & \textbf{416.44}($\pm$84.62) \\
 & \textit{Overlap Ratio $\downarrow$} & 0.684($\pm$0.097) & \textbf{0.023}($\pm$0.101) & 0.113($\pm$0.212) & 0.081($\pm$0.183) & 0.048($\pm$0.137) \\
  & \textit{Success rate $\uparrow$} & 98 & 45 & 82 & 73 & \textbf{100} \\
\midrule
\multirow{3}{*}{4} & \textit{Trajectory Length. $\downarrow$} & 417.92($\pm$89.76) & 427.30($\pm$86.46) & 416.50($\pm$80.35) & 433.03($\pm$96.89) & \textbf{357.5}($\pm$67.07) \\
 & \textit{90$\%$ Coverage $\downarrow$} & 346.50($\pm$80.11) & 358.56($\pm$83.64) & 329.75($\pm$81.73) & 318.46($\pm$95.15) & \textbf{294.21}($\pm$62.45) \\
 & \textit{Overlap Ratio $\downarrow$} & 0.693($\pm$0.096) & \textbf{0.075}($\pm$0.143) & 0.246($\pm$0.214) & 0.191($\pm$0.187) & 0.170($\pm$0.169) \\
& \textit{Success rate $\uparrow$} & 99 & 95 & \textbf{100} & 92 & \textbf{100} \\
\midrule
\multirow{3}{*}{8} & \textit{Trajectory Length. $\downarrow$} & 319.57($\pm$73.66) & 311.21($\pm$74.89) & 318.99($\pm$59.40) & 333.73($\pm$109.30) & \textbf{279.35}($\pm$50.63) \\
 & \textit{90$\%$ Coverage $\downarrow$} & 266.04($\pm$69.99) & 279.76($\pm$65.51) & 252.08($\pm$56.19) & 256.40($\pm$67.84) & \textbf{231.54}($\pm$46.40) \\
 & \textit{Overlap Ratio $\downarrow$} & 0.671($\pm$0.109) & \textbf{0.172}($\pm$0.198) & 0.401($\pm$0.199) & 0.306($\pm$0.177) & 0.316($\pm$0.174) \\
 & \textit{Success rate $\uparrow$} & 99 & \textbf{100} & 98 & 98 & \textbf{100} \\
\bottomrule
\end{tabular}%
}
\vspace{-0.2cm}
\end{table*}

\begin{table*}[tb]
\caption{\textbf{Adaptability to different sensor's FoV.} All tests are conducted with 4 agents and sensor range $d_s = 10$m.}
\vspace{-0.2cm}
\label{table:2}
\centering
\resizebox{\textwidth}{!}{%
\tiny
\aboverulesep=0.2mm \belowrulesep=0.2mm
\begin{tabular}{c|c|c|c|c|c|c}
\toprule
FoV & Metrics & Nearest & MMPF & NBVP & Learnt-Greedy & MARVEL \\
\midrule
\multirow{3}{*}{90$^{\circ}$} & \textit{Trajectory Length. $\downarrow$} & 414.64($\pm$82.28) & 472.92($\pm$87.65) & 455.28($\pm$82.03) & 511.77($\pm$103.13) & \textbf{378.99}($\pm$57.61) \\
 & \textit{90$\%$ Coverage $\downarrow$} & 347.46($\pm$84.07) & 388.45($\pm$80.47) & 362.53($\pm$85.70) & 366.97($\pm$102.75) & \textbf{313.67}($\pm$54.57) \\
 & \textit{Overlap Ratio $\downarrow$} & 0.594($\pm$0.125) & \textbf{0.059}($\pm$0.130) & 0.203($\pm$0.199) & 0.180($\pm$0.181) & 0.147($\pm$0.155) \\
 & \textit{Success rate $\uparrow$} & 97 & 90 & 98 & 94 & \textbf{100} \\
\midrule
\multirow{3}{*}{180$^{\circ}$} & \textit{Trajectory Length. $\downarrow$} & 422.80($\pm$108.04) & 389.08($\pm$89.53) & 380.55($\pm$84.31) & 424.74($\pm$105.14) & \textbf{328.36}($\pm$61.31) \\
 & \textit{90$\%$ Coverage $\downarrow$} & 348.05($\pm$95.57) & 320.36($\pm$70.86) & 320.88($\pm$83.17) & 315.65($\pm$83.23) & \textbf{273.18}($\pm$59.95) \\
 & \textit{Overlap Ratio $\downarrow$} & 0.807($\pm$0.124) & \textbf{0.098}($\pm$0.175) & 0.321($\pm$0.228) & 0.216($\pm$0.189) & 0.207($\pm$0.183) \\
 & \textit{Success rate $\uparrow$} & 99 & 96 & 99 & 97 & \textbf{100} \\
\bottomrule
\end{tabular}%
}
\vspace{-0.5cm}
\end{table*}

\begin{figure}
    \centering
    \includegraphics[width=0.45\textwidth]{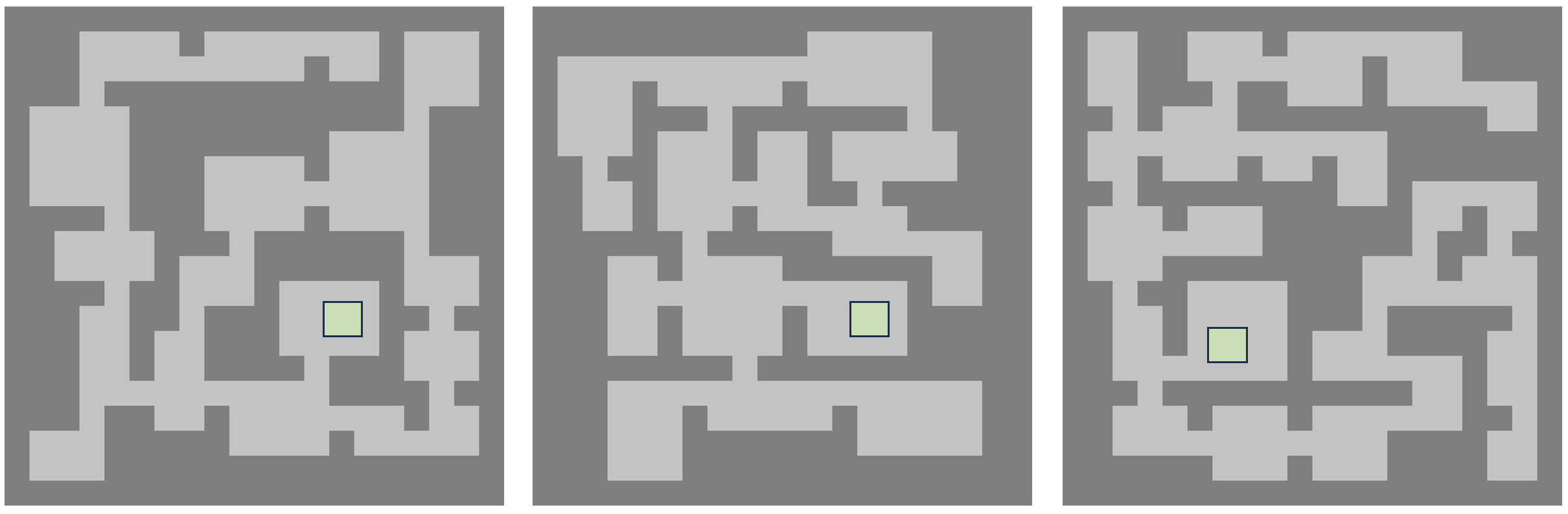}
    \caption{\textbf{Train/Test environments.} Green box: the starting region, which is kept constant for all evaluations conducted. }
    \label{fig:test_train_envs}
    \vspace{-0.55cm}
\end{figure}

\textbf{Reward structure:}
We designed a reward function to optimize the objective in ~\eqref{eq:obj}, defined as $r_i = r_o + a \cdot r_h + r_t + r_f$. This function incorporates:  $r_o$ , the reward based on the number of observable frontiers from the new viewpoint; $r_h$, the heading reward, calculated as the cosine of the angle deviation from the A* path  towards the nearest frontier, scaled by a factor aa (set to 0.3); $r_t$, a team-shared reward for the total number of frontiers observed by the team, where $r_o$ and $r_t$ are normalized by the maximum observable frontiers within the field of view; and $r_f$, a reward of +10 for task completion when the cumulative utility of nodes reaches 0. This reward structure encourages agents to explore new areas effectively and prioritize task completion while ensuring smooth trajectory planning. 

\subsection{Network Architecture}
Our policy network (Fig. \ref{fig:model}) utilizes an encoder-decoder architecture with stacked masked attention. The encoder refines node features in $G'_t$, which, combined with the robot's current location features, are fed to the decoder to generate waypoint-heading policies.  

\textbf{Encoder:} 
Our encoder leverages a sequence of masked self-attention layers to enhance the feature representations of each node by integrating information from other nodes within the graph. We begin by applying a feed-forward layer to transform the node properties in $V'_t$ into d-dimensional node features $h^n_1$. We apply 1D convolutions to $F_t$ that encodes the frontier distribution for each node individually to get frontier features $h^n_2$. This vector captures the node-specific frontier data, and extract local patterns and relationships within the node’s frontier distribution. The resulting node features $h^n_1$ and frontier features $h^n_2$ are concatenated, then projected back to $d$-dimensional space via a linear layer, to obtain node features  $h^n$. These features are then processed through $N$ stacked self-attention layers (with $N$=6 in our implementation), where each layer’s input is the output of the previous one (i.e., $h^q = h^{k,v} = h^n$). We apply a graph’s mask $M_t$, computed from the adjacency matrix of $E_t$, to ensures that each node only attends to features of its neighboring nodes (i.e., attention weigh,t $w_{ij} = 0, \ \forall (v_i, v_j) \notin E_t)$. While attention is restricted to neighboring nodes at each layer, nodes still gather non-neighboring node information by progressively aggregating features across the stacked self-attention layers. Unlike unmasked self-attention, our structured approach enhances path-finding performance by iteratively refining features through local connections. This enables the encoder to transform $h^n$ into $\tilde{h}^n$, where node representations capture broader graph dependencies influenced by $M_t$ and the attention depth $N$. 

\textbf{Policy decoder:} The decoder determines the final policy using the enhanced node features $\tilde{h}^n$. From these features, we select the node at the robot’s current position as the query: $h^q = h^c = \tilde{h}^n_{\psi_i(t)}$. The current node features $h^c$, along with neighboring node features from $\tilde{h}^n_e$, are input into an attention layer. This process refines the output $\tilde{h}^c$ by incorporating information from neighboring nodes. To improve the decoder’s awareness of the local environment, we compute two orientation-specific features for each $v_j$: We first discretize the heading space into 36 uniform bins, where each bin represents a distinct heading. Then we compute a binary feature vector based on the heading space for each $a_q$ of $v_j$, indicating whether the heading falls within the FoV centered on $a_q$. The second is another 36-bin vector that tracks the previously explored headings of $v_j$. This data is then processed through feed-forward layers to produce the corresponding embedding, ${h}^h_a$ and ${h}^h_e$. Since $\tilde{h}^n$ pertains solely to waypoints and does not include heading information, we replicate each neighboring node feature $x$ times, to generate joint action pairs. Then, we concatenate the expanded  $\tilde{h}^n$ with ${h}^h_a$ and ${h}^h_e$ and then projected back to d-dimensional space via a feed-forward layer to get heading enhanced neighboring features $\tilde{h}^n_e$. We then apply a single-head attention mechanism with $\tilde{h}^c$ and its enhanced neighboring features $\tilde{h}^n_e$, using the resulting attention scores directly as the final output policy $\pi_\theta = (a_{i,t} \mid o_{i,t}) = w_{i,j}$. This policy guides the agent in selecting the next viewpoint $(v_j, a_q)$. Its design removes the constraint of a fixed policy size, allowing dynamic adaptation to varying numbers of neighboring nodes.

\textbf{Critic decoder:} 
We developed our critic network, based on \cite{wangviper_2024}, to selectively extract pertinent information for each agent by incorporating other agents' actions. The critic network $Q_\phi$, parameterized by $\phi$, utilizes the same encoder structure as the policy network $\pi_\theta$ but features a distinct decoder. During training, the critic receives full ground-truth information, such as the entire map or graph (see Fig.~\ref{fig:model}), enhancing training stability through \textit{privileged learning}.  Using the enhanced node features $\tilde{h}^n$ from the encoder, we retrieve the features of other agents, $\tilde{h}^n_{\psi_{-i}(t)}$, and their corresponding features at the next timestep, $\tilde{h}^n_{\psi_{-i}(t+1)}$, based on their current actions $a_t$. These key-value pairs augment the current node feature $h^c$ with other agents' actions, aiding credit assignment.. The enhanced current node features $\tilde{h}^c$ are combined with enhanced neighboring features $\tilde{h}^n_e$, computed in the same manner as in the policy decoder. Finally, these concatenated features are mapped to state-action values $Q_\phi(s_t, a_t)$ using a feed-forward layer. The networks are trained using soft actor-critic (SAC) algorithm \cite{christodoulou2019sac}.

\textbf{Training details:} 
We use a training dataset of 5,663 randomly generated large-scale maps (see Fig. \ref{fig:test_train_envs}) and an additional 100 unseen maps for testing. Each environment measures $90$m by $90$m with 4 agents, a sensor range of $d_s=10$m, and a 120$^\circ$ FoV. Training parameters include a maximum episode length of 128 steps, a discount factor of $\gamma=1$, a batch size of 256, and an episode buffer size of 10,000. Training starts after accumulating over 2,000 steps in the buffer. We set the target entropy to 0.01 $\cdot \log(k)$, perform one iteration per training step after each episode, and use the Adam optimizer with a learning rate of $10^{-5}$ for both policy and critic networks. The target critic network updates every 256 steps. Complete code and model are available at \url{https://github.com/marmotlab/MARVEL}.

\section{Experiments}
\subsection{Comparison Analysis}
Many prior studies evaluate exploration planners in a limited number of scenarios, often fewer than 10. However, we observed that performance can vary significantly across different scenarios, necessitating evaluation across a broad range of environments. Therefore, we test our model and baselines in 100 unseen environments. We compare MARVEL with the following baselines: \textbf{Nearest} \cite{Yamauchi_nearest_1998}, which selects the nearest frontier as the global goal using a merged map of all agents; \textbf{MMPF} \cite{yu_apf_2021}, which computes a potential field based on the explored map and agent positions, guiding agents to frontiers via steepest descent, with a resistance force to prevent redundancy. To decrease the computation complexity, frontiers will be gathered into multiple clusters. In each iteration, agents will select the smallest potential cluster as their target; \textbf{NBVP} \cite{Bircher_2016}, which samples and evaluates trajectories based on information gain, executing the most optimal one; and \textbf{Learnt-Greedy} \cite{cao_large_24}, which employs deep reinforcement learning with graph-based attention to select the next sampling location, adapted for multi-agent systems with greedy headings based on observable frontiers.

We present the average and variance of \textit{trajectory length} of complete exploration (99$\%$ coverage), \textit{90$\%$ coverage}, \textit{overlap ratio} and \textit{success rate} in Table \ref{table:1}. The \textit{overlap ratio} quantifies the shared sensing area among agents per timestep, indicating collaborative efficiency. \textit{90$\%$ coverage} refers to the maximum distance required for agents to explore 90$\%$ of the environment, with lower values indicating higher efficiency. Success rate measures the percentage of tests where agents achieve 99$\%$ coverage within 128 timesteps. Our results show that MARVEL outperforms all baselines in average trajectory length, surpassing the best baseline, NBVP, by 14.2$\%$ with four agents. Additionally, MARVEL achieved the highest success rate across all test scenarios, attaining a perfect 100$\%$ success rate in all tests, which highlights its robustness.  MARVEL also demonstrates superior stability with the lowest variance in trajectory metrics, highlighting its strong generalization. Computation times are not included as the baselines were not optimized for efficiency, focusing instead on their core implementations.

In terms of \textit{overlap ratio}, MARVEL ranks second in most tests, with MMPF emerging as the top performer. This is expected, as MMPF’s design naturally disperses agents using strong repulsive forces when agents are in close proximity, encouraging them to explore independently. Despite this, MARVEL demonstrates superior multi-agent coordination, excelling in trajectory metrics. In contrast, the Nearest method perform poorly, likely due to their short-sighted strategies, which is more prominent in our large-scale environments. Also, MMPF’s focus on local exploration further hinders effective agent coordination. NBVP faces challenges in larger environments due to its sampling-based approach, and while increasing its sampling rate could improve performance, it comes with high computational costs. In comparison, MARVEL achieves consistent computation times under $0.2$s per decision across all scenarios.

\begin{table}[t]
    \tiny
    \captionsetup{skip=1pt}
    \caption{\textbf{Adaptability to different sensor ranges.} All tests are conducted with 4 agents and 120$^{\circ}$ FoV.}
    \aboverulesep=0.3mm \belowrulesep=0.3mm
    \begin{minipage}[t]{0.48\textwidth} 
        \makebox[0pt][l]{ 
        \resizebox{\textwidth}{!}{%
        \begin{tabular}{c|c|c}
        \toprule
        Metrics & 12m & 15m \\
        \midrule
        \textit{Trajectory Length $\downarrow$} & 299.57($\pm$60.33) & 258.35($\pm$54.96) \\
        \textit{90$\%$ Coverage $\downarrow$} & 247.93($\pm$58.7) & 212.85($\pm$49.5) \\
        \textit{Overlap Ratio $\downarrow$} & 0.223($\pm$0.188) & 0.208($\pm$0.190) \\
        \bottomrule
        \end{tabular}%
        }}
        \vspace{-0.6cm}
        \label{table:3}
    \end{minipage}
\end{table}

Upon closer inspection, we noticed that learnt-greedy methods delivers faster coverage of 90$\%$ of the environment than nearest and MMPF, indicating the strong performance of graph-based attention networks in learning spatial relationship for exploration tasks. However, this method struggles with small scattered frontiers, produced due to suboptimal viewpoint orientations, leading to repeated revisits and longer trajectories for task completion. As a result, learnt-greedy do not perform well when comparing the trajectory length for task completion. In contrast, MARVEL's attention-based neural network, with intelligent fusion of frontiers and orientation information, enhances the environmental understanding and provides efficient reasoning about large partial maps, enabling non-myopic viewpoints planning.

\subsection{Adaptability to different sensor configurations}
We extensively evaluated our models across various FoV configurations, as detailed in Table \ref{table:2}, to understand the impact of sensor parameters on performance. By avoiding retraining with different FoV settings, we isolated performance variations to sensor differences. Our method consistently outperformed all baselines in trajectory metrics, demonstrating robust and efficient exploration planning under varying FoV conditions. Additionally, we tested the model with different sensor ranges, as shown in Table \ref{table:3}. Originally trained with $d_s = 10$m, the model successfully utilized extended ranges of $d_s = 12$m and $15$m, achieving notable performance gains, and maintaining a 100$\%$ success rate. This adaptability without retraining highlights the model’s resilience to sensor changes and its ability to effectively use additional sensory data, making it suitable for diverse environments.

\subsection{Experimental Validation}
\begin{figure}
    \centering
    \includegraphics[width=0.4\textwidth]{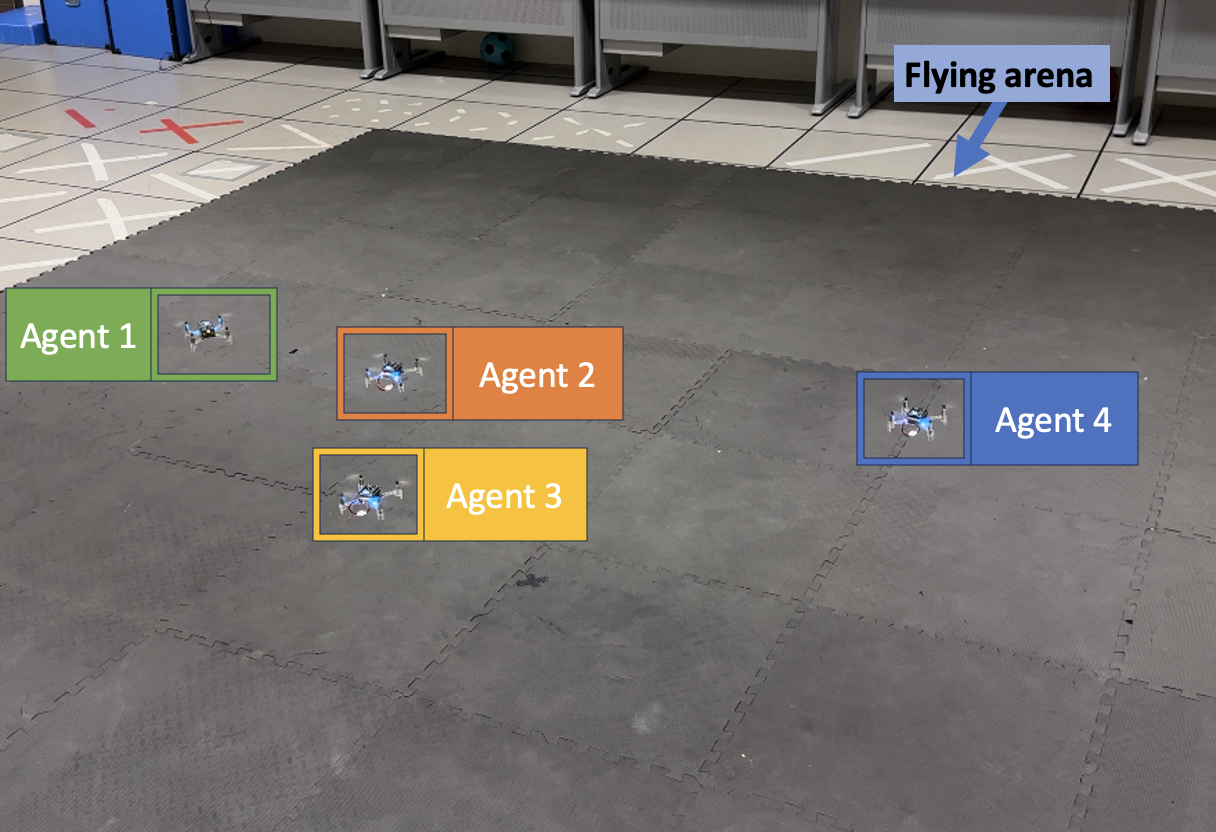}
    \caption{\textbf{Experimental validation on four nano drones.} The grey mat demarcates the 4m by 4m flying arena.}
    \label{fig:drone}
    \vspace{-0.65cm}
\end{figure}

To assess MARVEL’s potential for real-world deployment, we conducted experiments with Crazyflie 2.1 drones in a $4$m by $4$m arena (see Fig. \ref{fig:drone}). Using the swarm manager from \cite{chiun2024star}, we simulated sensor coverage based on the drones’ real-time positions, enabling live virtual mapping tests to be conducted. These experiments confirmed that MARVEL can be successfully implemented on physical hardware and adapts well to the motion dynamics of the drones.

\section{Conclusion}
In this work, we introduce a neural framework tailored for multi-robot exploration with constrained field-of-view (FoV) sensors in large-scale environments. Our approach employs a graph attention network, with intelligent fusion of frontier and orientation features, to enhance environmental understanding and allow agents to produce non-myopic decisions. In particular, MARVEL handles the extensive action space of viewpoint planning using an information-driven action pruning method.  Our evaluations show that MARVEL outperforms existing state-of-the-art multi-agent exploration planners, adapts well to different team sizes and sensor configurations (i.e., FoV and sensor range), and performs reliably across diverse environments. Additionally, MARVEL has been successfully validated on real drones, demonstrating its potential for deployment on actual robots.

While MARVEL excels in 2D indoor settings, we acknowledge the added complexity of 3D environments with intricate obstacles. To address this, we plan to extend MARVEL to handle full 3D action spaces by integrating height information and associated FoV changes, thereby enhancing its application on platforms such as drones for indoor mapping. There, to ensure efficient 3D exploration, we plan to employ a sparse graph representation to manage computational demands effectively.

\bibliographystyle{IEEEtran}
\bibliography{citations.bib}

\end{document}